\documentclass[runningheads]{llncs}
\usepackage{epsfig}
\usepackage{graphicx}
\usepackage{caption}
\usepackage{amsmath}
\usepackage{amssymb}
\usepackage{color}
\usepackage{booktabs}
\usepackage{multirow}
\usepackage{diagbox}
\usepackage{subfig}
\usepackage{enumitem}

\usepackage{authblk}

\usepackage{lipsum}

\newcommand\blfootnote[1]{%
  \begingroup
  \renewcommand\thefootnote{}\footnote{#1}%
  \addtocounter{footnote}{-1}%
  \endgroup
}

\definecolor{NKcolor}{RGB}{100, 0, 200}

\begin{document}
\setlength{\abovedisplayskip}{1pt}
\setlength{\belowdisplayskip}{1pt}
% ===================================================================
% title
% ===================================================================
\title{Semi-Supervised and Task-Driven Data Augmentation}
%\title{Task Driven Data Augmentation}
%\author[1]{Alice Smith}
%\author[2]{Bob Jones}
%\affil[1]{Department of Mathematics, University X}
%\affil[2]{Department of Biology, University Y}

\author{Krishna Chaitanya\inst{1}, Neerav Karani\inst{1}, Christian Baumgartner\inst{1}, Olivio Donati\inst{2}, Anton Becker\inst{2}, Ender Konukoglu\inst{1}}
\institute{Computer Vision Lab, ETH Zurich \and
  University Hospital of Zurich, Zurich, Switzerland}
%Computer Vision Lab, ETH Zurich, Zurich, Switzerland}
%Princeton University, Princeton NJ 08544, USA \and
%Springer Heidelberg, Tiergartenstr. 17, 69121 Heidelberg, Germany
%\email{{lncs@springer.com}}
%\email{abc@uni-heidelberg.de}
%
\maketitle              % typeset the header of the contribution
%

%\renewcommand\Authands{ and }
% ===================================================================
% abstract
% ===================================================================
\begin{abstract}
Supervised deep learning methods for segmentation require large amounts of labelled training data, without which they are prone to overfitting, not generalizing well to unseen images.
%In practical scenarios
In practice, obtaining a large number of annotations from clinical experts is expensive and time-consuming.
One way to address scarcity of annotated examples is data augmentation using random spatial and intensity transformations.
%Recently, several generative models based on Generative Adversarial Networks (GANs) have been proposed to synthesize realistic training examples, complementing the random augmentation.
Recently, it has been proposed to use generative models to synthesize realistic training examples, complementing the random augmentation.
So far, these methods have yielded limited gains over the random augmentation.
However, there is potential to improve the approach by (i) explicitly modeling deformation fields (non-affine spatial transformation) and intensity transformations and (ii) leveraging unlabelled data during the generative process.
%Current methods using generative modelling do not consider one or more of the following factors into their framework: they do not (i) leverage unlabelled data, (ii) explicitly model non-affine transformations and (iii) explicitly model intensity transformations.
%%%%%%%%% re-phrase %%%%%%%%%
%In this work, we propose a novel task-driven data augmentation approach where the generated images from the available labelled images are transformed via a deformation field and an additive intensity mask.
%such that the generated data is conducive to the segmentation task.
With this motivation, we propose a novel task-driven data augmentation method where to synthesize new training examples, a generative network explicitly models and applies deformation fields and additive intensity masks on existing labelled data, modeling shape and intensity variations, respectively. 
%%%%%%%%%
%why is this formulation important
%why deformation - to model shape variations
%why intensity - to model intensity variations
%The deformation field and additive intensity mask are intended to model the shape variations and intensity variations present in the dataset.
%In this work, we propose a novel task-driven data augmentation approach that models augmented images from the available labelled images where they are transformed via a deformation field and an additive intensity mask.
Crucially, the generative model is optimized to be conducive to the task, in this case segmentation, and constrained to match the distribution of images observed from labelled and unlabelled samples.
%Here, we constrain the generated images to match the distribution of all labelled and unlabelled images, and the generated image-label pairs to be conducive for the learning of the task defined e.g. segmentation task. 
%This formulation also provides appropriately transformed segmentation masks for the augmented images with exact correspondence between the image and label, unlike earlier approaches.
Furthermore, explicit modeling of deformation fields allow synthesizing segmentation masks and images in exact correspondence by simply applying the generated transformation to an input image and the corresponding annotation.
Our experiments on cardiac magnetic resonance images (MRI) showed that, for the task of segmentation in small training data scenarios, the proposed method substantially outperforms conventional augmentation techniques.
%as well as semi-supervised learning methods. 
%Our experimental results also showed the semi-supervised learning methods we analyzed did not yield additional performance increase on top of traditional augmentation.
%that aim to span diversity of intensity characteristics commonly found in Magnetic Resonance imaging (MRI). 
\end{abstract}
% ===================================================================
% intro
% ===================================================================
\section{Introduction}~\label{sec:intro}
% ======================================
% training with a small training dataset is hard
% ======================================
\blfootnote{This article has been accepted at the 26th international conference on Information Processing in Medical Imaging (IPMI) 2019.}
Precise segmentation of anatomical
%and pathological
structures is crucial for several clinical applications.
Recent advances in deep neural networks yielded automatic segmentation algorithms with unprecedented accuracy.
However, such methods heavily rely on large annotated training datasets.
In this work, we consider the problem of medical image segmentation in the setting of small training datasets.
%\add[kc]{test sentence}
%\remove[editor]{removed text}
%\change[editor]{removed text}{added text}
%such as diagnosis of pathologies and treatment planning

%%%DL does not work if we have limited data & obtaining annotations is costly
%%%Intensity problems as well with MR imaging
% ======================================
% training with a small training dataset is hard
% ======================================
\vspace{0.1cm} \noindent Let us first consider the question: \textit{why is a large training dataset necessary for the success of deep learning methods?}
One hypothesis is that a large training dataset exposes a neural network to sufficient variations in factors, such as shape, intensity and texture, thereby allowing it to learn a robust image to segmentation mask mapping.
In medical images, such variations may arise from subject specific shape differences in anatomy or lesions.
Image intensity and contrast characteristics may differ substantially according to the image acquisition protocol or even between scanners for the same acquisition protocol.
When the training dataset is small, deep learning methods are susceptible to faring poorly on unseen test images either due to not identifying such variations or because the test images appear to have been drawn from a distribution different to the training images.

\vspace{0.1cm} \noindent We conjecture that one way to train a segmentation network on a small training dataset more robustly could be to incorporate into the training dataset, intensity and anatomical shape variations observed from a large pool of unlabelled images.
Specifically, we propose to generate synthetic image-label pairs by learning generative models of deformation fields and intensity transformations that map the available labelled training images to the distribution of the entire pool of available images, including labelled as well as unlabelled.
%When learning a segmentation task, the training data set should be large enough to allow a neural network to infer the target relationship between images and segmentation masks under the variations in different factors including shape and intensity characteristics of the target and surrounding structure. Algorithms trained on small data sets may not be exposed to a sufficient amount of variation and consequently be limited in their ability to generalize outside of the observed data. Variations may either arise from difference in the patient anatomy or pathology, or from differences in the image acquisition. The latter is particularly relevant to magnetic resonance (MR) imaging where intensity and contrast properties are dependent on the scanner and acquisition protocol which may change from patient to patient. 
% ======================================
% task driven data augmentation
% ======================================
Additionally, we explicitly encourage the synthesized image-label pairs to be conducive to the task at hand.
% \noindent \textit{motivate task-driven data augmentation...}
%We carry out extensive evaluation of the proposed method that indicates that it substantially outperforms existing data augmentation as well as semi-supervised learning techniques for segmentation of cardiac MRIs.
We carried out extensive evaluation of the proposed method, in which the method showed substantial improvements over existing data augmentation as well as semi-supervised learning techniques for segmentation of cardiac MRIs.
%instead of semi-supervised

% ======================================
% related work
% ======================================
\vspace{0.1cm} \noindent \textbf{Related work}:
Due to the high cost of obtaining large amount of expert annotations, robust training of machine learning methods in the small training dataset setting has been widely studied in the literature.
Focusing on the methods that are most relevant to the proposed method, we broadly classify the related works into two categories:
%deep learning methods in the small training dataset setting has been widely studied in the literature.
%The related works can be broadly classified into two categories:

% ======================================
% data augmentation
% ======================================
\vspace{0.1cm} \noindent \textbf{Data augmentation} is a technique wherein the training dataset is enlarged with artificially synthesized image-label pairs.
The main idea is to transform training images in such a way that the corresponding labels are either unchanged or get transformed in the same way.
Some commonly used data augmentation methods are affine transformations~\cite{cirecsan2011high} (such as translation, rotation, scaling, flipping, cropping, etc.) and random elastic deformations~\cite{simard2003best,ronneberger2015u}.
Leveraging recent advances in generative image modelling~\cite{goodfellow2014generative}, several works proposed to map randomly sampled vectors from a simple distribution to realistic image-label pairs as augmented data for medical image segmentation problems~\cite{bowles2018gan,costa2018end,shin2018medical}. Such methods are typically trained on already labelled data, with the objective of interpolating within the training dataset.
In an alternative direction, \cite{zhang2017mixup} proposed to synthesize data for augmentation by simply linearly interpolating the available images and the corresponding labels. Surprisingly, despite employing clearly unrealistic images, this method led to substantial improvements in medical image segmentation~\cite{eaton2018improving} when the available training dataset is very small.
None of these data augmentation methods use unlabelled images that may be more readily available and all of them, except for those based on generative models, are hand-crafted rather than optimized based on data.%heuristically designed.
\vspace{0.1cm} \noindent \textbf{Semi-supervised learning} (SSL) methods are another class of techniques that are suitable in the setting of learning with small labelled training datasets.
The main idea of these methods is to regularize the learning process by employing unlabelled images.
%, relatively large numbers of which are comparatively cheaper to obtain. 
Approaches based on self-training~\cite{bai2017semi} alternately train a network with labeled images, estimate labels for the unlabelled images using the network and update the network with both the available true image-label pairs and the estimated labels for the unlabelled images.
%Graph-based methods leverage unlabeled data to regularize the classifier for medical image segmentation~\cite{su2016interactive}.
\cite{zhang2017deep} propose a SSL method based on adversarial learning, where the joint distribution of unlabelled image-estimated labels pairs is matched to that of the true labelled images-label pairs. Interestingly,~\cite{oliver2018realistic} show that many SSL methods fail to provide substantial gains over the supervised baseline that is trained with data augmentation and regularization.
\vspace{0.1cm} \noindent \textbf{Weakly-supervised learning} tackles the issue of expensive pixel-wise annotations by training on weaker labels, such as scribbles~\cite{can2018scribble} and image-wide labels~\cite{andermatt2018pathology}.
Finally, other regularization methods that do not necessarily leverage unlabelled images may also aid in preventing over-fitting to small training datasets.
\vspace{-0.1cm}
\section{Methods}
\vspace{-0.1cm}
% ===================================================
% supervised learning
% ===================================================
In a supervised learning setup, an objective function $L_S(\{X_L,Y_L\})$ that measures discrepancy between ground truth labels, $Y_L$, and predictions of a network $S$ on training images, $X_L$, is minimized with respect to a set of learnable parameters $w_S$ of the network, i.e.
\begin{equation}
    \min_{w_{S}} \Big(L_S(\{X_L,Y_L\})\Big)
    \label{eq:sup_obj}
\end{equation}
% ===================================================
% supervised learning + data augmentation
% ===================================================
\noindent When data augmentation is employed in the supervised learning setup, Eq.~\ref{eq:sup_plus_aug_obj} is minimized with respect to $w_S$.
\begin{equation}
    \min_{w_{S}} \Big(L_S(\{X_L,Y_L\} \cup \{X_G,Y_G\})\Big)
    \label{eq:sup_plus_aug_obj}
\end{equation}
Here, $X_G$ and $Y_G$ refer to generated images and labels obtained by affine or elastic transformations of $X_L$, $Y_L$ or by using methods such as Mixup~\cite{zhang2017mixup}.
The set $\left\{ \{X_L,Y_L\} \cup \{X_G, Y_G\}\right\}$ is referred to as the augmented training set.
% ===================================================
% supervised learning + data augmentation adverserial approach
% ===================================================
In augmentation methods based on generative models~\cite{bowles2018gan}, the parameters of $S$ are still optimized according to Eq.~\ref{eq:sup_plus_aug_obj}, but the generative process for $X_G$ and $Y_G$ involves two other networks: a generator $G$ and a discriminator $D$.
The corresponding parameters $w_G$ and $w_D$ are estimated according to the generative adversarial learning (GAN)~\cite{goodfellow2014generative} framework by optimizing:
\begin{equation}
    \min_{w_G} \max_{w_D} \mathbb{E}_{x,y\sim p(x_L,y_L)}[\log D(x, y)] + \mathbb{E}_{z\sim p_z(z)}[\log(1 - D(G(z))] 
    \label{eq:gan}
\end{equation}
$G$ takes as input a vector $z$ sampled from a known distribution $p_z(z)$ and maps that to a \{$X_G$, $Y_G$\} pair.
$D$ is optimized to distinguish between outputs of $G$ and real \{$X_L$, $Y_L$\} pairs, while $G$ is optimized to generate \{$X_G$, $Y_G$\} pairs such that $D$ responds to them similarly as to \{$X_L$, $Y_L$\}.
Thus, $\{X_G, Y_G\}$ are encouraged to be ``realistic'' in the sense that they cannot be distinguished by $D$.
%The loss used for optimizing $w_G$ may be thought of as a regularizer, $L_{G,reg}$, on the generative model.

% ===================================================
% proposed approach
% ===================================================
\vspace{-0.1cm}
\subsection{Semi-Supervised and Task-Driven Data Augmentation}
%We propose to generate the augmentation image-label pairs (\{$X_G$, $Y_G$\}) using a different GAN framework that enforces them to be conducive to the task as hand.
Instead of solving the optimization given in Eq.~\ref{eq:gan} for generating the augmentation image-label pairs \{$X_G$, $Y_G$\}, we propose solving Eq.~\ref{eq:sup_plus_aug_task_driven_obj}:
\begin{equation}
    \min_{w_G}\Big(\min_{w_{S}} L_S(\{X_L,Y_L\} \cup \{X_G,Y_G\}) +
    L_{reg, w_G}(\{X_L\}\cup\{X_{UL}\})\Big)
    %L_{reg,}(\{X_L, Y_L\}\cup\{X_{UL}\}\cup\{X_G,Y_G\})\Big)
    \label{eq:sup_plus_aug_task_driven_obj}
\end{equation}
This incorporates two ideas.
The first term dictates that \{$X_G$, $Y_G$\} be such that they are beneficial for minimizing the segmentation loss $L_S$.
Secondly, note that $L_{reg, w_G}$ depends not only on the labelled images \{$X_L$\} (as in Eq.~\ref{eq:gan}), but also on the unlabelled images \{$X_{UL}$\}.
It is a regularization term based on an adversarial loss, which incorporates information about the image distribution that can be extracted from both \{$X_L$\} and \{$X_{UL}$\}.
%$L_{reg}$ is a regularization term based on an adversarial loss, which incorporates the information on the image distribution that can be extracted from the observed samples, including both labelled and unlabelled examples, $X_{UL}$.
%\noindent Further, in order to involve the unlabelled images in the data augmentation, we model the generative process to map the labelled images to the unlabelled images.
%\noindent Further, in order to incorporate the information on the image distribution that can be extracted from the observed samples, we model the generative process to map between image pairs in the observed set.
This is achieved by synthesizing $\{X_G\}$ as $G_C(\{X_L\})$, where $G_C$ denotes a conditional generative model.
$G_C$ is modelled in two different ways: one for deformation field (non-affine spatial transformations) and one for intensity transformations.
In both cases, the formulation is such that as a certain labelled image is mapped to an augmentation image, the mapping to obtain the corresponding augmentation label readily follows.
%This is achieved by synthesizing $\{X_G, Y_G\}$ using two conditional generative models: one for spatial transformations and one for intensity transformations, as described below.
%To synthesize $\{X_G, Y_G\}$, we use two conditional generator models, one for spatial transformations and one for intensity transformations, both conditioned on an input image coming from the labelled set.
%This is done in two steps using a conditional spatial transformation GAN and a conditional intensity transformation GAN.
%We propose a generative modelling based approach using conditional GAN (cGAN). We train two separate conditional GANs to model two primary factors of variation in : (i) shape using deformation fields to obtain non-affine transformations and (ii) intensity using additive intensity field to obtain intensity transformations. We have a U-Net~\cite{ronneberger2015u} attached to each cGAN that is trained for the segmentation task as illustrated in the Fig.~\ref{fig:gans}.

% ===================================================
% Deformation field generator
% ===================================================
\vspace{-0.1cm}
\subsubsection{Deformation Field Generator}:
The deformation field generator, $G_C$ = $G_V$, is trained to create samples from the distribution of deformation fields that can potentially map elements from $\{X_L\}$ to those in the combined set $\{X_L\} \cup \{X_{UL}\}$.
%The objective of this module (Fig.~\ref{fig:gans}(a)) is to model the distribution of \textit{deformation fields} that can potentially map the set of labelled images $\{X_L\}$ to that of unlabelled images $\{X_{UL}\}$.
$G_V$ takes as input an image from $\{X_L\}$ and a vector $z$, sampled from a unit Gaussian distribution, and outputs a dense per-pixel deformation field, $\textbf{v}$.
%The input image is warped using bilinear interpolation according to $\textbf{v}$ to produce $X_{G,V}$.
The input image and its corresponding label (in 1-hot encoding) are warped using bilinear interpolation according to $\textbf{v}$ to produce $X_{G,V}$ and $Y_{G,V}$ respectively.
%Accordingly, the segmentation mask represented as 1-hot encoding is also warped using bi-linear interpolation according to $\textbf{v}$ to produce $Y_{G,V}$. 
%The segmentation mask corresponding to the input image is also warped using nearest neighbour interpolation according to $\textbf{v}$ to produce $Y_{G,V}$. 

% ===================================================
% Additive intensity field generator
% ===================================================
\vspace{-0.1cm}
\subsubsection{Additive Intensity Field Generator}:
The intensity field generator, $G_C$ = $G_I$, is trained to draw random samples from the distribution of additive intensity fields that can potentially map elements from $\{X_L\}$ to those in $\{X_L\} \cup \{X_{UL}\}$.
%labelled images to the combined set of labelled and unlabelled images. %, similar to $G_V$.
$G_I$, takes as input an element of $\{X_L\}$ and a noise vector and outputs an intensity mask, $\Delta I$.
$\Delta I$ is added to the input image to give the transformed image $X_{G,I}$, while its segmentation mask $Y_{G,I}$ remains the same as that of the input image.

\setlength{\belowcaptionskip}{-10pt}
\begin{figure}[t!]
    \subfloat[Deformation field cGAN]{\includegraphics[width=0.5\textwidth]{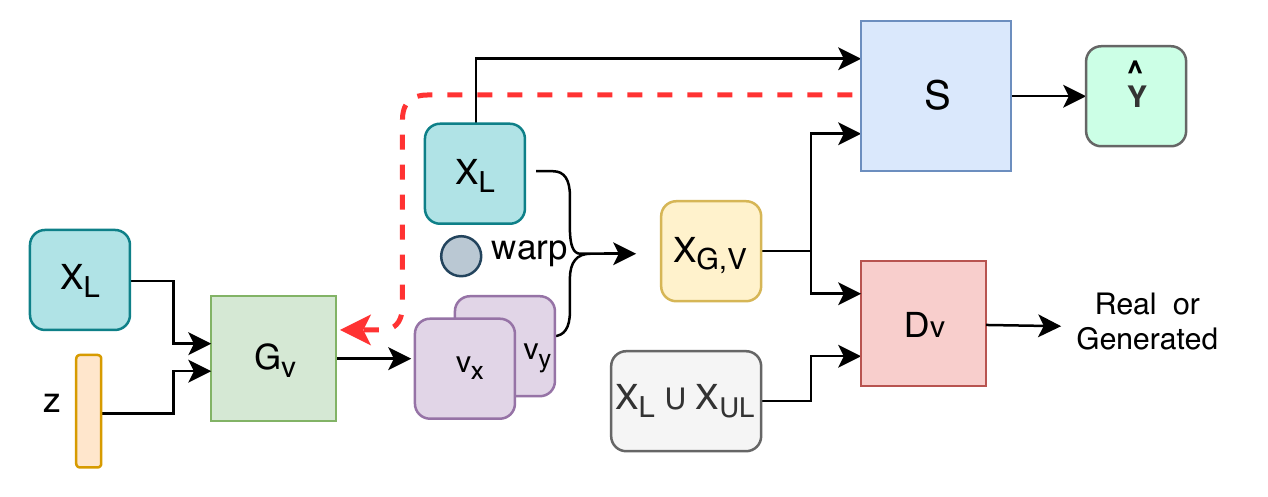}}
    \subfloat[Additive Intensity field cGAN]{\includegraphics[width=0.5\textwidth]{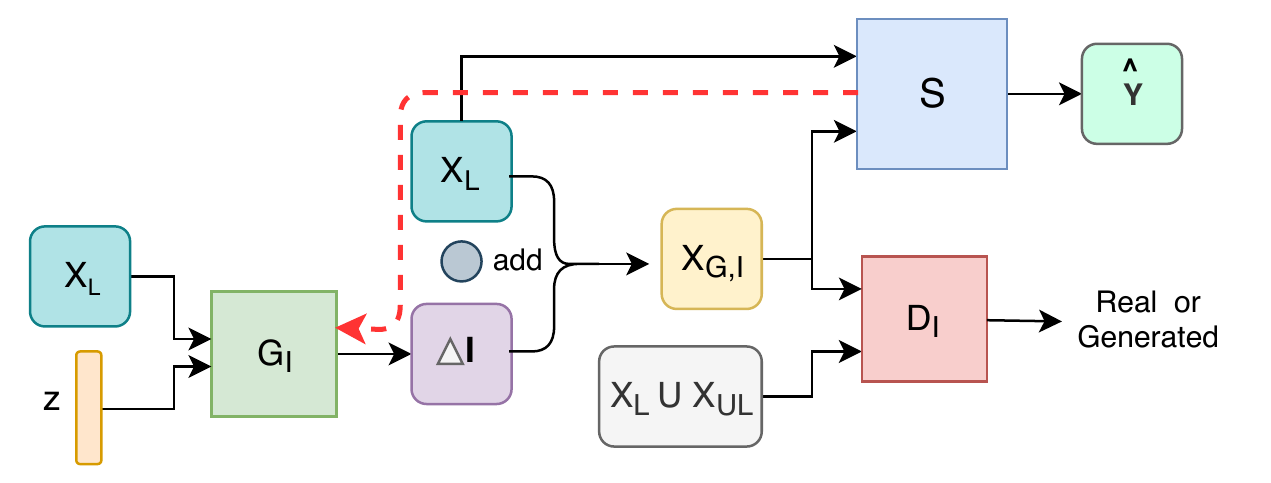}}
    \caption{Modules for task-driven and semi-supervised data augmentation.}
    \label{fig:gans}
\end{figure}

% ===================================================
% Regularization term
% ===================================================
\vspace{-0.1cm}
\subsubsection{Regularization Loss}:
For both the conditional generators, the regularization term, $L_{reg, w_G}$, in Eq.~\ref{eq:sup_plus_aug_task_driven_obj} is formulated as in Eq.~\ref{eq:gan_G_C}.
The corresponding discriminator networks $D_C$ are trained to minimize the usual adversarial objective (Eq.~\ref{eq:gan_D_C}).
\begin{equation}
    L_{reg, w_{G_{C}}} = 
    \lambda_{adv}\mathbb{E}_{z\sim p_z(z)}[\log(1 - D_{C}(G_{C}(z,X_L)))] + \lambda_{big}L_{G_C,big}
    \label{eq:gan_G_C}
\end{equation}
\begin{equation}
    L_{D_C} = 
    - \mathbb{E}_{x\sim p(x)}[\log D_{C}(x)]
    - \mathbb{E}_{z\sim p_z(z)}[\log(1 - D_{C}(G_{C}(z,X_L)))] 
    \label{eq:gan_D_C}
\end{equation}

\noindent The generated images are obtained as
$G_{V}(z,X_L) = \textbf{v} \circ X_L$ and
$G_{I}(z,X_L) = \Delta I + X_L$,
where $\circ$ denotes a bilinear warping operation.
In our experiments, we observe that with only the adversarial loss term in $L_{reg, w_{G_{C}}}$, the generators tend to create only the identity mapping.
So, we introduce the $L_{G_C,big}$ term to incentivize non-trivial transformations.
We formulate $L_{G_V,big}$ and $L_{G_I,big}$ as $- ||\textbf{v}||_{1}$ and $- ||\Delta I||_{1}$ respectively.

% ===================================================
% Optimization
% ===================================================
\vspace{-0.1cm}
\subsubsection{Optimization Sequence}:
The method starts by learning the optimal data augmentation for the segmentation task.
Thus, all networks $S$, $G_C$ and $D_C$ are optimized according to Eq.~\ref{eq:sup_plus_aug_task_driven_obj}.
The generative models for the deformation fields and the intensity fields are trained separately.
Once this is complete, both $G_V$ and $G_I$ are fixed and the parameters of $S$ are re-initialized.
Now, $S$ is trained again according to Eq.~\ref{eq:sup_plus_aug_obj}, using the original labelled training data $\{X_L, Y_L\}$ and augmentation data $\{X_G, Y_G\}$ generated using the trained $G_V$ or $G_I$ or both.

\vspace{-0.1cm}
\section{Dataset and Network details}
\vspace{-0.1cm}
% =============================================
% subsection: Dataset
% =============================================
\subsection{Dataset Details}~\label{sec:dataset}
%\textbf{Cardiac MRI}:
We used a publicly available dataset hosted as part of MICCAI'17 ACDC challenge~\cite{bernard2018deep}~\footnote{https://www.creatis.insa-lyon.fr/Challenge/acdc}.
It comprises of short-axis cardiac cine-MRIs of 100 subjects from 5 groups - 20 normal controls and 20 each with 4 different cardiac abnormalities.
%(abnormal right ventricles, dilated cardiomyopathy, previous myocardial infarction and hypertrophic cardiomyopathy).
The in-plane and through-plane resolutions of the images range from 0.70x0.70mm to 1.92x1.92mm and 5mm to 10mm respectively. Expert annotations are provided for left ventricle (LV), myocardiam (Myo) and right ventricle (RV) for both end-systole (ES) and end-diastole (ED) phases of each subject. For our experiments, we only used the ES images.
%It comprises of short-axis cardiac cine-MRIs of 100 subjects acquired with 1.5T and 3T scanners.

% =============================================
% subsection: Pre-processing
% =============================================
\vspace{-0.1cm}
\subsection{Pre-processing}
\vspace{-0.1cm}
We apply the following pre-processing steps to all images of the dataset:
(i) bias correction using N4~\cite{tustison2010n4itk} algorithm,
(ii) normalization of each 3d image by linearly re-scaling the intensities as: $(x-x_2)/(x_{98}-x_2)$, where $x_2$ and $x_{98}$ are the $2^{nd}$ and $98^{th}$ percentile in the bias corrected 3d image,
(iii) re-sample each slice of each 3d image and the corresponding labels to an in-plane resolution of 1.367x1.367mm using bi-linear and nearest neighbour interpolation respectively and crop or pad them to a fixed size of 224x224.
%We used only end-systole (ES) phase images of ACDC dataset for all the experiments. All the cardiac MR images were re-sampled to standard in-plane resolution of 1.367x1.367mm using bi-linear interpolation and labels using nearest neighbours and cropped to fixed image size of 224x224. Images were bias corrected using N4~\cite{tustison2010n4itk} with a threshold value of 0.001. Each 3D image is normalized using min-max normalization where the maximum and minimum value corresponds to the $98^{th}$ percentile and $2^{nd}$ percentile respectively to get rid of the outliers.

% =============================================
% subsection: Network Architectures
% =============================================
\vspace{-0.1cm}
\subsection{Network Architectures}
\vspace{-0.1cm}
There are three types of networks in the proposed method (see Fig.~\ref{fig:gans}): a segmentation network $S$, a generator network $G$ and a discriminator network $D$.
In this sub-section, we describe their architectures.
Expect for the last layer of $G$, the same architecture is used for the $G_V$ and $G_I$ networks used for modelling both the deformation fields and the intensity transformations.
%The cGAN for modelling deformation and additive intensity fields comprise of generator and discriminator networks and a U-Net each as illustrated in Fig.~\ref{fig:gans}. 

% ==================================
% generator
% ==================================
\vspace{0.1cm} \noindent \textbf{Generator}:
$G$ takes as input an image from $\{X_L\}$ and a noise vector $z$ of dimension 100, which are both first passed through separate sub-networks, $G_{subnet,X}$ and $G_{subnet,z}$. 
$G_{subnet,X}$, consists of 2 convolutional layers,
while $G_{subnet,z}$, consists of a fully-connected layer, followed by reshaping of the output, followed by 5 convolutional layers, interleaved with bilinear upsampling layers.
The outputs of the two sub-networks are of the same dimensions.
They are concatenated and passed through a common sub-network, $G_{subnet,common}$, consisting of 4 convolutional layers, the last of which is different for $G_V$ and $G_I$.
The final convolutional layer for $G_V$ outputs two feature maps corresponding to the 2-dimensional deformation field $\textbf{v}$, while that for $G_I$ outputs a single feature map corresponding to the intensity mask $\Delta I$.
The final layer of $G_I$ employs the tanh activation to cap the range of the intensity mask.
All other layers use the ReLU activation.
All convolutional layers have 3x3 kernels except for the final ones in both $G_V$ and $G_I$ and are followed by batch-normalization layers before the activation.

%\textbf{Generator}: The conditional generator is fed with an input labeled image and a 1D random vector of size 100. The generator has a fully connected (dense) layer over these 100 random numbers and the result is reshaped into down-sampled image dimensions. Then, a set of 4 successive upsampling and convolution layers are applied to get the resultant output in the shape of image dimensions. The input image is convolved with two convolution layers to output features which are concatenated to the output features obtained from 1D random vector. This concatenated feature map is input to three more convolution operations to get the penultimate convolution layer($F_{l}$).
%The generator is fed with an input of a sampled labeled image and a 1D vector of 100 random numbers drawn from a gaussian distribution with zero mean and standard deviation of one. 
%For deformation field modelling, the final convolution is performed on $F_{l}$ with a 3x3 kernel to get the per-pixel flow vector.
%For intensity field, the final convolution is applied on $F_{l}$ with a 1x1 kernel and the result is passed through a tanh activation layer to get the additive intensity field.
%The upsamplings are done using bilinear interpolation.
%The convolutions use ReLU activation and batch norm~\cite{ioffe2015batch} is applied on the layers except the final layers for stable training of the cGANs.

% ==================================
% discriminator
% ==================================
\vspace{0.1cm} \noindent \textbf{Discriminator}:
%$D$ has a similar architecture to the DCGAN~\cite{radford2015unsupervised} architecture.
$D$ consists of 5 convolutional layers with kernel size of 5x5 and stride 2.
The convolutions are followed by batch normalization layers and leaky ReLU activations with the negative slope of the leak set to 0.2.
After the convolutional layers, the output is reshaped and passed through 3 fully-connected layers, with the final layer having an output size of 2.
%softmax classifier which classifies an input image is a generated or real image.

% ==================================
% segmentation network
% ==================================
\vspace{0.1cm} \noindent \textbf{Segmentation Network}:
We use a U-net~\cite{ronneberger2015u} like architecture for $S$.
It has an encoding and a decoding path.
In the encoder, there are 4 convolutional blocks, each consisting of 2 3x3 convolutions, followed by a max-pooling layer.
The decoder consists of 4 convolutional blocks, each made of a concatenation with the corresponding features of the encoder, followed by 2 3x3 convolutions, followed by bi-linear upsampling with factor 2.
Batch normalization and ReLU activation are employed in all layers, except the last one.
\vspace{-0.1cm}
\subsection{Training Details}
\vspace{-0.1cm}
Weighted cross-entropy is used as the segmentation loss, $L_S$.
%For the segmentation loss, $L_S$, we use the weighted cross-entropy loss.
We empirically set the weights of the 4 output labels to 0.1 (background) and 0.3 (each of the 3 foreground labels).
The background loss is considered while learning the augmentations, but not while learning the segmentation task alone.
% hyperparameters: stopping criterion, batch size, loss function
\noindent We empirically set $\lambda_{adv}$ and $\lambda_{big}$ to 1 and $10^{-3}$ respectively.
The batch size is set to 20 and each training is run for 10000 iterations.
The model parameters that provide the best dice score on the validation set are chosen for evaluation.
Adam optimizer is used for all networks with an initial learning rate of $10^{-3}$, $\beta_1=0.9$ and $\beta_2=0.999$.

% ===================================================================
% section: Experiments
% ===================================================================
\vspace{-0.2cm}
\section{Experiments}~\label{sec:experiment_setup}
%\vspace{-0.1cm}
%0. Why use only 2D over 3D images for training ?
%A. GPU memory limitations for higher batch, slice thickness varies and non isotropic images need different kernel size of 3D filter in axial direction.
%We use 2D networks for all the experiments. The reasons for this choice is: (i) due to the dataset having higher slice thickness in the axial direction than the resolution in the in-plane direction e.g. the in-plane resolution ranges from 0.7mm to 1.92mm and slice thickness ranges from 5 to 10mm. There are roughly 7-10 axial slices for each patient of ACDC data. Hence, we cannot use isotropic kernels to capture same information from these 3D images. Also, (ii) GPU memory limitations hinder the training of 3D networks with higher batch size.
%0.1 Use traditional 2D U-Net for training baseline
%1.0 No of images for training, validation and test sets
%1.1 No of cross validations
% ================================
% number of images in training / validation / test
% ================================
We divide the dataset into test ($X_{ts}$), validation ($X_{vl}$), labelled training ($X_{L}$) and unlabelled training ($X_{UL}$) sets which consist of 20, 2, N\textsubscript{L} and 25 3d images respectively. As we are interested in the few labelled training images scenario, we run all our experiments in two settings: with N\textsubscript{L} set to 1 and 3.
% ================================
% selection of these subsets
% ================================
$X_{ts}$, $X_{vl}$ and $X_{UL}$ are selected randomly a-priori and fixed for all experiments. $X_{ts}$ and $X_{UL}$ are chosen such that they consist of equal number of images from each group (see Sec.~\ref{sec:dataset}) of the dataset.
A separate set of 10 images (2 from each group), $X_{L,total}$, is selected randomly.
Each experiment is run 5 times with $X_{L}$ as N\textsubscript{L} images randomly selected from $X_{L,total}$.
When N\textsubscript{L} is 3, it is ensured that the images in $X_{L}$ come from different groups.
Further, each of the 5 runs with different $X_{L}$ is run thrice in order to account for variations in convergence of the networks.
Thus, overall, we have 15 runs for each experiment.
%The dataset is divided into training, validation and test sets. The training set is further divided into labeled and unlabeled training sets. The test set consists of 20 3d images and is not involved in any training phase. The validation set consists of 2 3d images. The unlabeled training data is a fixed set of images for all experiments which is set to 25 3d images. The labeled set comprises of a pool of annotated training images. We experiment with 2 setups with different number of labelled training images, $N_{labelled} = 1,3$.% i.e. 1, 3 training images used for training of each setup.
% ================================
% number of runs for each exp setting
% ================================
%In each experiment, we sample 5 different sets of training images randomly from the defined pool of training images. Each set of selected training images is run 3 times to account for variations in convergence of the network. Thus, overall we have 15 experiment runs for each setup.

% mention details of loss functions in sec. 'Training'
%The model used for computing the dice score of test subjects is the one that yields best dice score on the validation set. The batch size used is 20. Weighted cross-entropy score is used as the loss for the training of the segmentation network. The experiments are listed into below subsections.

\vspace{0.1cm} \noindent The following experiments were done thrice for each choice of $X_{tr,L}$:
\begin{itemize}[leftmargin=*]

    %1,1 No data augmentation 
    \item \textbf{No data augmentation (Aug\textsubscript{none})}: $S$ is trained without data augmentation.
    
    %1,2 Affine data augmentation 
    \item \textbf{Affine data augmentation(Aug\textsubscript{A})}: $S$ is trained with data augmentation comprising of affine transformations.
    These consist of
    rotation (randomly chosen between -15deg and +15deg),
    scaling (with a factor randomly chosen uniformly between 0.9 and 1.1),
    %cropping () and
    another possible rotation that is multiple of 45deg (angle=45deg*N where N is randomly chosen between 0 to 8),
    and flipping along x-axis.
    For each slice in a batch, a random number between 0 and 5 is uniformly sampled and accordingly, either the slice is left as it is or is transformed by one of the 4 stated transformations.
    % In the first experiment, we train two base 2D U-Net segmentation networks with the limited training images : (i) one without any data augmentation and (ii) second with data augmentation comprising of affine transformations. Data augmentation ($X_{a}$) used for this experiment consist popular transformations such as rotation, scaling, cropping, flipping, etc. of the training images. This network trained with data augmentation acts as a \textit{baseline} for other methods. The network is trained using a learning rate of $10^{-3}$.
    
    \vspace{0.1cm} \noindent All the following data augmentation methods, each training batch (batch size$=$bs) is first applied affine transformations as explained above. The batch used for training consists of half of these images along with bs/2 augmentation images obtained according to the particular augmentation method.

    %3.1 Elastic deformations
    \item \textbf{Random elastic deformations (Aug\textsubscript{A,RD})}:
    Elastic augmentations are modelled as in~\cite{ronneberger2015u}, where a deformation field is created by sampling each element of a 3x3x2 matrix from a Gaussian distribution with mean 0 and standard deviation 10 and upscaling it to the image dimensions using bi-cubic interpolation.
    %The smooth elastic deformations applied on images are obtained as stated in~\cite{ronneberger2015u}, where the mean and standard deviation are chosen as 0 and 10.
    
    %3.2 Random contrast and brightness augmentation
    \item \textbf{Random contrast and brightness fluctuations}~\cite{hong2017convolutional,perez2018data} \textbf{(Aug\textsubscript{A,RI})}:
    This comprises of an image contrast adjustment step: $x = (x - \Bar{x}) * c + \Bar{x}$, followed by a brightness adjustment step: $x = x + b$.
    We sample c and b uniformly in [0.8,1.2] and [-0.1,0.1] respectively.
    %The image contrast is applied using a scaling factor that is sampled from an uniform distribution of [0.8,1.2]. The image brightness step includes adding a constant random number to all the pixels which is sampled from the range of [-0.1 to 0.1].
    %In this experiment, we apply random contrast and brightness augmentation~\cite{perez2018data,hong2017convolutional} on top of traditional augmentation.

    %4 Deformation field cGAN
    \item \textbf{Deformation field transformations (Aug\textsubscript{A,GD})}:
    Augmentation data is generated from the trained deformation field generator $G_V$.
    %In this experiment, we use our proposed data augmentation to generate non-affine transformed images ($X_{v}$) as stated in Fig.~\ref{fig:gans}(a). These generated images and original labelled images are used for training the U-Net where the generated images assist in yielding higher segmentation performance.

    %5 Intensity field cGAN
    \item \textbf{Intensity field transformations (Aug\textsubscript{A,GI})}:
    Augmentation data is generated from the trained intensity field generator $G_I$.
    %In this experiment, we use our proposed data augmentation to generate intensity transformed images ($X_{i}$) as stated in Fig.~\ref{fig:gans}(b). These generated images and original labelled images are used for training the U-Net where the generated images also assist in improving the segmentation performance.

    %6. Both GANs augmented data 
    \item \textbf{Both deformation and intensity field transformations (Aug\textsubscript{A,GD,GI})}: In this experiment, we sample data from $G_V$ and $G_I$ to obtain transformed images $X_{V}$ and $X_{I}$ respectively.
    We also get an additional set of images which contain both deformation and intensity transformations $X_{VI}$.
    These are obtained by conditioning $G_I$ on spatially transformed images $X_{V}$.
    The augmentation data comprises of all these images $\{X_{V},X_{I},X_{VI}\}$.

    %7. Mix up  
    \item \textbf{MixUp}~\cite{zhang2017mixup} \textbf{(Aug\textsubscript{A,Mixup})}:
    Augmentation data ($\{X_G, Y_G\}$) is generated using the original annotated images $X_L$ and their linear combinations using the Mixup formulation as stated in Eq.~\ref{eq:mixup_eqn}.
    \begin{equation}
        {X_G} = \lambda X_{Li} + (1-\lambda) X_{Lj},\hspace{0.2cm}
        {Y_G} = \lambda Y_{Li} + (1-\lambda) Y_{Lj}
        \label{eq:mixup_eqn}
    \end{equation}
    where $\lambda$ is sampled from beta distribution Beta$(\alpha,\alpha)$ with $\alpha \in (0,\infty)$ and $\lambda \in [0,1)$ which controls the ratio to mix the image-label pairs $(X_{Li},Y_{Li})$, $(X_{Lj},Y_{Lj})$ selected randomly from the set of labelled training images.

    %8. Mixup + our method
    \item \textbf{Mixup over deformation and intensity field transformations \\ (Aug\textsubscript{A,GD,GI,Mixup})}:
    Mixup is applied over different pairs of available images: original data ($X_L$), their affine transformations and the images generated using deformation and intensity field generators $\{X_{V},X_{I},X_{VI}\}$.

    %9. Semi-supervised Learning 
    \item \textbf{Adversarial Training (Adv Tr)}:
    Here, we investigate the benefit of the method proposed in~\cite{zhang2017deep} on our dataset (explained in Sec.~\ref{sec:intro}), in both supervised (SL)~\cite{luc2016semantic} and semi-supervised (SSL)~\cite{zhang2017deep} settings.
    %In this experiment, we train the U-Net that serves as a generator along with a discriminator. For supervised learning (SL) and semi-supervised learning setting, we use the method as stated in~\cite{luc2016semantic} and as in~\cite{zhang2017deep} respectively.
    
\end{itemize}

\noindent \textbf{Evaluation} : The segmentation performance of each method is evaluated using the Dice similarity coefficient (DSC) over 20 test subjects for three foreground structures: left ventricle (LV), myocardiam (Myo) and right ventricle (RV).
%All the experiments include affine data augmentation except first experiment where it is explicitly trained without data augmentation.
%DSC is used to measure the overlap between the predicted segmentation mask and the ground truth.

% =======================================================
% =======================================================
\vspace{-0.1cm}
\section{Results and Discussion}
\begin{figure}[t!]
    \hspace{1.0cm}
    (a) \hspace{1.0cm}
    (b) \hspace{1.0cm}
    (c) \hspace{1.0cm}
    (d) \hspace{1.0cm}
    (e) \hspace{1.0cm}
    (f) \hspace{1.0cm}
    (g) \\
    \includegraphics[width=1.0\textwidth,trim={0cm 1cm 0cm 0.7cm},clip]{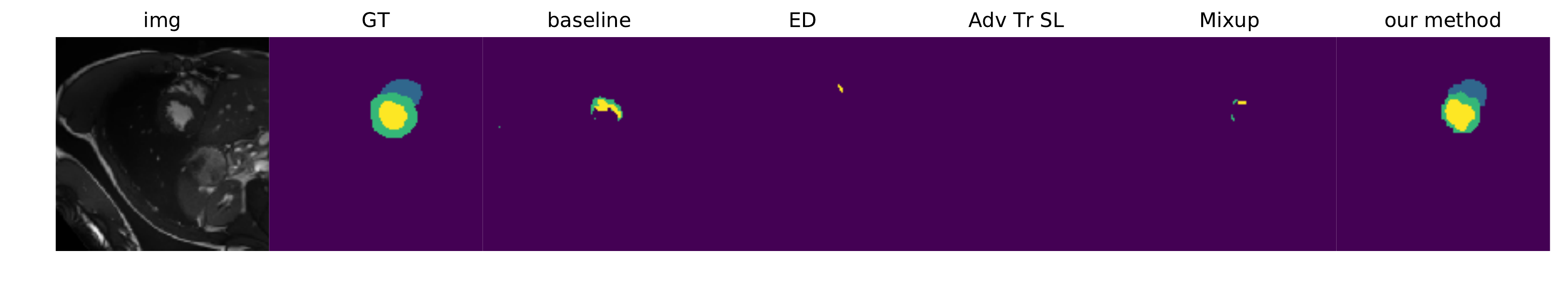}\\    
    \includegraphics[width=1.0\textwidth,trim={0cm 1cm 0cm 0.4cm},clip]{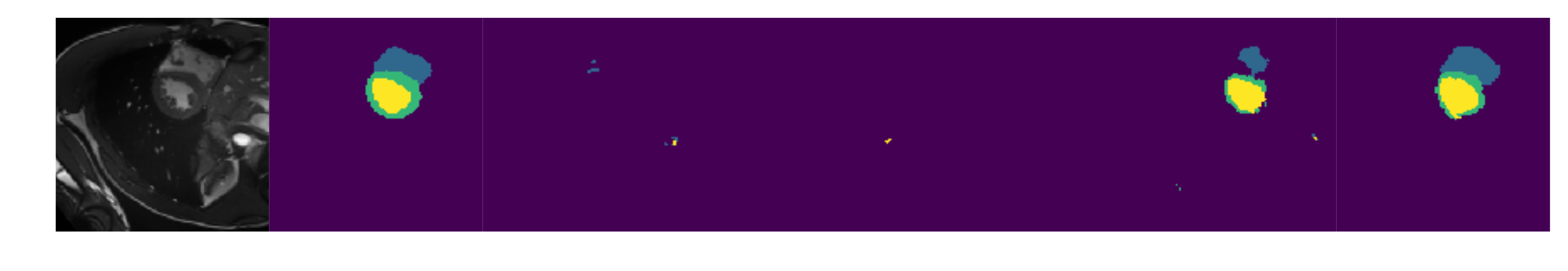}\\
    \includegraphics[width=1.0\textwidth,trim={0cm 1cm 0cm 0.4cm},clip]{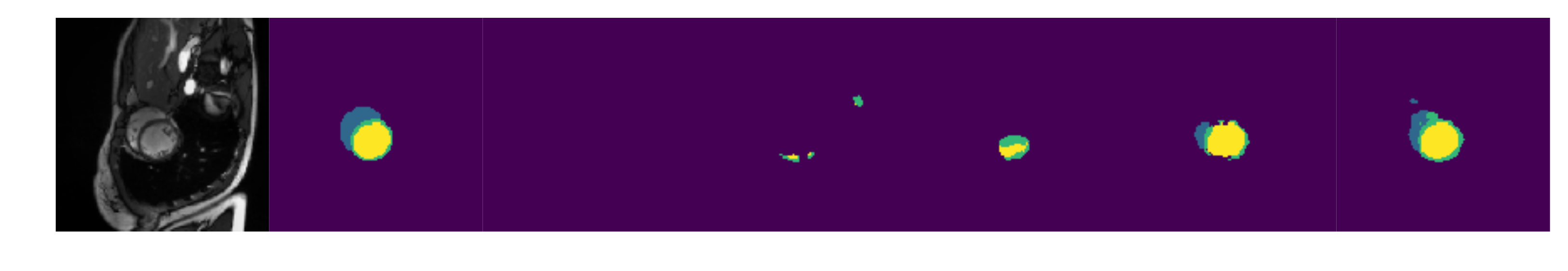}\\
    \includegraphics[width=1.0\textwidth,trim={0cm 1cm 0cm 0.4cm},clip]{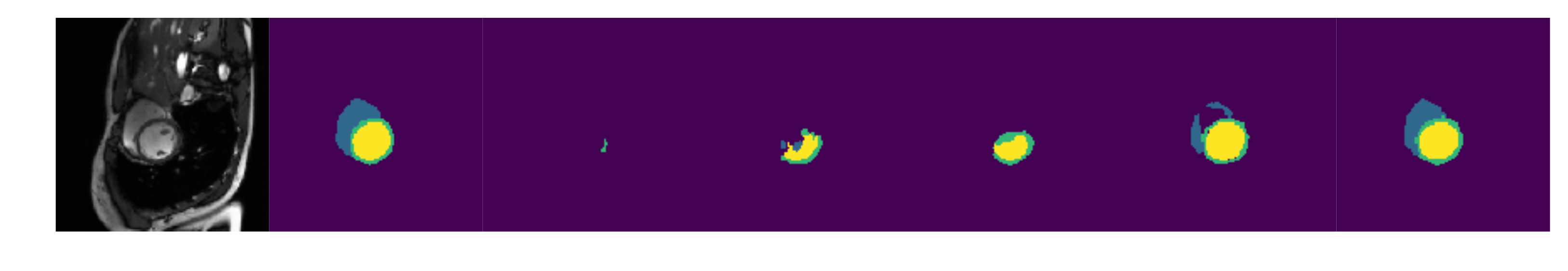}\\
    \includegraphics[width=1.0\textwidth,trim={0cm 1cm 0cm 0.4cm},clip]{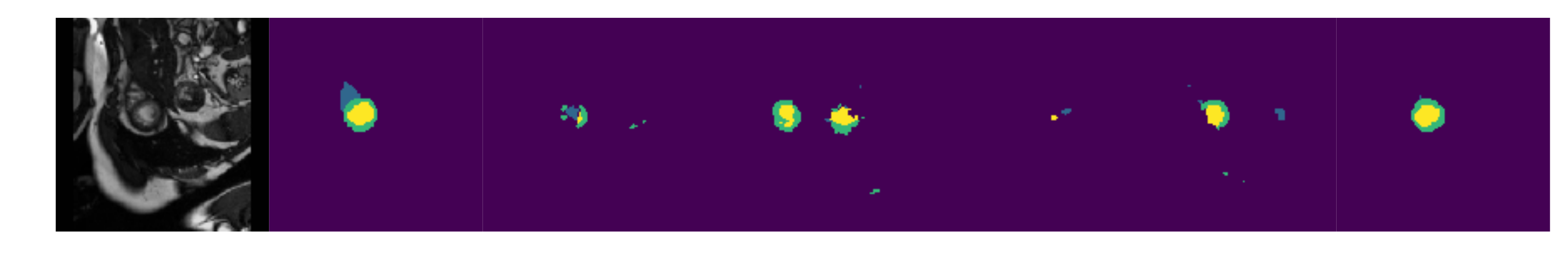}\\
    \includegraphics[width=1.0\textwidth,trim={0cm 1cm 0cm 0.4cm},clip]{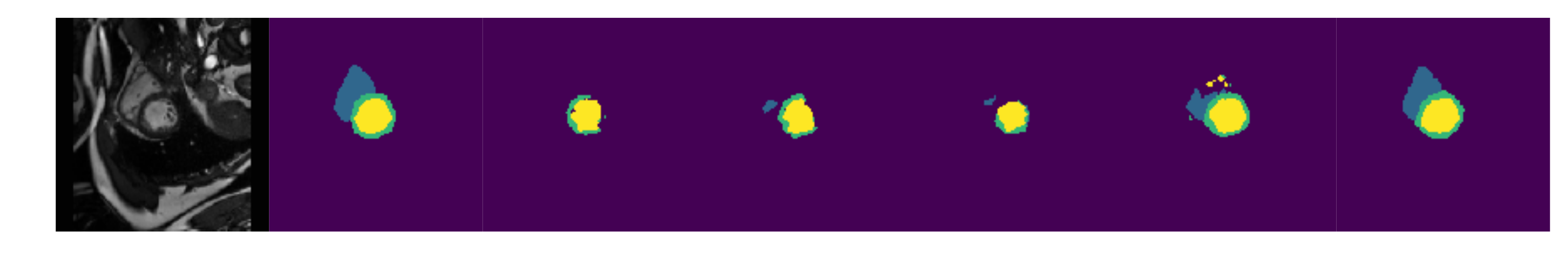}\\
    %\caption{Segmentation Results of proposed method vs earlier approaches}
    \caption{Qualitative comparision of the proposed method with other approaches: (a) input image, (b) ground truth, (c) Aug\textsubscript{A}, (d) Aug\textsubscript{A,RD}, (e) Adv Tr SL~\cite{luc2016semantic}, (f) Aug\textsubscript{A,Mixup}~\cite{zhang2017mixup}, (g) Aug\textsubscript{A,GD,GI}}
    \vspace{-0.1cm}
    \label{fig:seg_results}
\end{figure}

\vspace{-0.1cm}
Table~\ref{table:acdc_results} presents quantitative results of our experiments.
The reported numbers are the mean dice scores over the 15 runs for each experiments as described in Sec.~\ref{sec:experiment_setup}.
It can be observed that the proposed method provides substantial improvements over other data augmentation methods as well as the semi-supervised adversarial learning method, especially in the case where only 1 3D volume is used for training.
The improvements can also be visually observed in Fig.~\ref{fig:seg_results}.
In the rest of this section, we discuss the results of specific experiments.

%affine data augmentation
\vspace{0.1cm} \noindent Perhaps unsurprisingly, the lowest performance occurs when neither data augmentation nor semi-supervised training is used.
Data augmentation with affine transformations already provides remarkable gains in performance.
Both random elastic deformations and random intensity fluctuations further improve accuracy.
%Similarly, augmentations with random intensity fluctuations also lead to an improvement over having no augmentations.
%We now checked if the proposed methods can provide improvements over such heuristic augmentations.

\begin{figure}[t!]
    \includegraphics[width=1.0\textwidth,trim={0cm 9.35cm 0cm 0.6cm},clip]{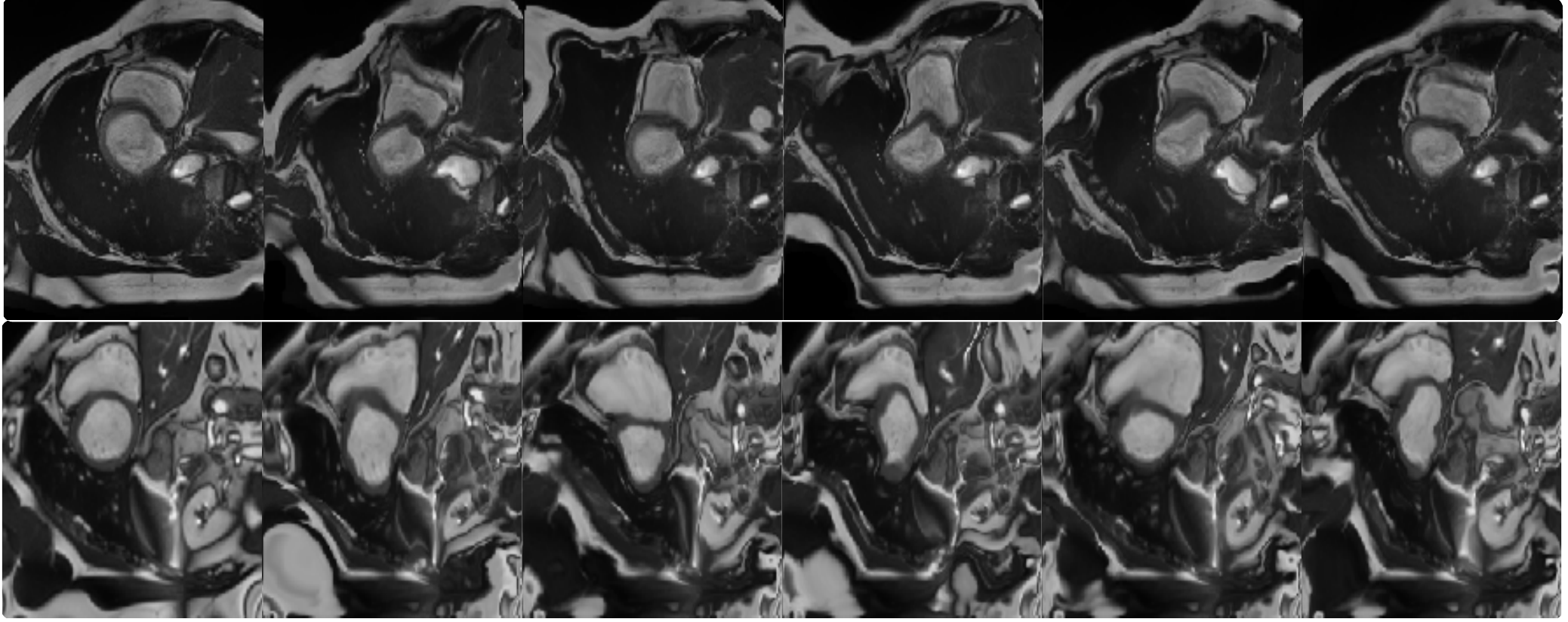}
    \\ Input Image \hspace{0.05cm} $|$ Images generated by $G_V$ $\rightarrow$ \\
    \includegraphics[width=1.0\textwidth,trim={0cm 0cm 0cm 9.6cm},clip]{images/geo_trans_imgs/geo_int_trans_imgs.pdf}
    \\ Input Image \hspace{0.05cm} $|$ Images generated by $G_I$ $\rightarrow$
    \caption{Generated augmentation images from the deformation field generator $G_V$ (top) and the intensity field generator $G_I$ (bottom).
    %Top image contains generated images with deformation fields and bottom image contains generated images with additive intensity fields applied
    }
    \vspace{-0.1cm}
    \label{fig:gen_geogan_imgs}
\end{figure}

% DFGAN improvements
\vspace{0.1cm} \noindent The proposed augmentations based on learned deformation fields improve performance as compared to random elastic augmentations. These results show the benefit of encouraging the deformations to span the geometric variations present in entire population (labelled as well as unlabelled images), while still generating images that are conducive to the training of the segmentation network. Some examples of the generated deformed images are shown in Fig.~\ref{fig:gen_geogan_imgs}. Interestingly, the anatomical shapes in these images are not always realistic. While this may appear to be counter-intuitive, perhaps preserving realistic shapes of anatomical structures in not essential to obtaining the best segmentation neural network.

% AIFGAN improvements
\vspace{0.1cm} \noindent Similar observations can be made about the proposed augmentations based on learned additive intensity masks as compared to random intensity fluctuations. Again, the improvements may be attributed to encouraging the intensity transformations to span the intensity statistics present in the population, while being beneficial for the segmentation task. Qualitatively, also as before, the generated intensity masks (Fig.~\ref{fig:gen_geogan_imgs}) do not necessarily lead to realistic images.

%DFGAN and AIFGAN together
\vspace{0.1cm} \noindent As both $G_V$ and $G_I$ are designed to capture different characteristics of the entire dataset, using both the augmentations together may be expected to provide a higher benefit than employing either one in isolation. Indeed, we observe a substantial improvement in dice scores with our experiments.

\vspace{0.1cm} \noindent As an additional experiment, we investigated the effect of excluding the regularization term from the training of the generators, $G_V$ and $G_I$ ($\lambda_{adv} = \lambda_{big} = 0$).
While the resulting augmentations still resulted in better performance than random deformations or intensity fluctuations, their benefits were lesser than that from the ones that were trained with the regularization.
This shows that although the adversarial loss does not ensure the generation of realistic images, it is still advantageous to include unlabelled images in the learning of the augmentations.
%The generalizability of this segmentation network is higher than the ones modeling individual factors since it observes both different shapes and intensity characteristics of the anatomical structures during training.
%We can also interpret that the transformations generated can map the limited labelled images into the space of unlabeled samples in the data manifold and thus, create diverse set of data points.

% Mixup
\vspace{0.1cm} \noindent Augmentations obtained from the Mixup~\cite{zhang2017mixup} method also lead to a substantial improvement in performance as compared to using affine transformations, random elastic transformations or random intensity fluctuations. Interestingly, this benefit also occurs despite the augmented images being not realistic looking at all. One reason for this behaviour might be that the Mixup augmentation method provides soft probability labels for the augmented images - such soft targets have been hypothesized to aid optimization by providing more task information per training sample~\cite{hinton2015distilling}. Even so, Mixup can only generate augmented images that are linear combinations of the available labelled images and it is not immediately clear how to extend this method to use unlabelled images.
% Mixup + our method
Finally, we see that Mixup provides a marginal improvement when applied over the original images together with the augmentations obtained from the trained generators $G_V$ and $G_I$.
This demonstrates the complementary benefits of the two approaches.
%For the last experiment, we apply Mixup over the generated augmented data from our method to benefit from both our proposed method and Mixup technique.
%Mixup can create a diverse set of samples since it can combine any pair of images from the generated $\{X_V,X_I,X_{VI}\}$ and original images $\{X_L\}$. These samples can cover much more space of the data manifold than either of the method independently. Hence, we get further improvements in dice scores as listed in Table~\ref{table:acdc_results}.

%Adv Training
\vspace{0.1cm} \noindent While semi-supervised adversarial learning provides improvement in performance as compared to training with no data augmentation, these benefits are only as much as those obtained with simple affine augmentation.
This observation seems to be in line with works such as~\cite{oliver2018realistic}.
%With adversarial training technique which has been quite popular among medical imaging communities in past few years, we observe small improvements over the baseline in the limited data setting.
%This can be due to the fact that in limited data scenarios, the segmentation network (generator) overfits to the data and training of the discriminator with few image-label pairs to estimate the joint distribution can lead to sub-optimal performance.

% ================================================================
% quantitative results
% ================================================================
\begin{table*}[!htb]
\setlength\belowcaptionskip{-15pt}
%\begin{tabular}{|p{4.25cm}|p{1cm}|p{1cm}|p{1cm}|p{1cm}|p{1cm}|p{1cm}|}
\begin{tabular}{|p{5cm}|p{1.1cm}|p{1.1cm}|p{1.1cm}|p{1.1cm}|p{1.1cm}|p{1.1cm}|}
\hline
\multirow{3}{*}{Method}                           & \multicolumn{6}{c|}{Number of 3D training volumes used}
\\ \cline{2-7} 
\multirow{2}{*} & \multicolumn{3}{c|}{1} & \multicolumn{3}{c|}{3}
\\ \cline{2-7}
%\diagbox{Methods}{No of training examples} & \multicolumn{3}{|c|}{1} & \multicolumn{3}{|c|}{3} & \multicolumn{3}{|c|}{5} \\ \cline{2-10} 
                                                    & RV & Myo & LV & RV & Myo & LV \\ \hline
Aug\textsubscript{none}                             & 0.259 & 0.291 & 0.446 & 0.589 & 0.631 & 0.805 \\ \hline
Aug\textsubscript{A}                              & 0.373 & 0.484 & 0.644 & 0.733 & 0.744 & 0.885 \\ \hline
Aug\textsubscript{A,RD}                   & 0.397 & 0.503 & 0.663 & 0.756 & 0.763 & 0.897 \\ \hline
Aug\textsubscript{A,GD}($\lambda_{adv}=1,\lambda_{big}=10^{-3}$)& $0.487^{\ast}$ & $0.560^{\ast}$ & $0.717^{\ast}$ & $0.782^{\ast}$ & $0.791^{\ast}$ & $0.908^{\ast}$ \\ \hline
Aug\textsubscript{A,GD}($\lambda_{adv}=0, \lambda_{big}=0$)& 0.394 & $0.531^{\ast}$ & $0.694^{\ast}$ & 0.756 & $0.776^{\ast}$ & $0.908^{\ast}$ \\ \hline
Aug\textsubscript{A,RI}                      & 0.429 & 0.554 & 0.742 & 0.744 & 0.759 & 0.896 \\ \hline
Aug\textsubscript{A,GI}($\lambda_{adv}=1,\lambda_{big}=10^{-3}$)& $0.517^{\dagger}$ & $0.579^{\dagger}$ & $0.773^{\dagger}$ & $0.803^{\dagger}$ & $0.791^{\dagger}$ & 0.912 \\ \hline
Aug\textsubscript{A,GI}($\lambda_{adv}=0, \lambda_{big}=0$)   & $0.500^{\dagger}$ & $0.583^{\dagger}$ & $0.765^{\dagger}$ & $0.766^{\dagger}$ & $0.748^{\dagger}$ & 0.893 \\ \hline
Aug\textsubscript{A,GD,GI}($\lambda_{adv}=1,\lambda_{big}=10^{-3}$)& \underline{$0.651^{\star}$} & \underline{$0.710^{\star}$} & \underline{$0.834^{\star}$} & \underline{$0.832^{\star}$} & \underline{$0.823^{\star}$} & \underline{$0.922^{\star}$} \\ \hline
Aug\textsubscript{A,Mixup}~\cite{zhang2017mixup}  & 0.581 & 0.599 & 0.774 & 0.818 & 0.791 & 0.915 \\ \hline
Aug\textsubscript{A,GD,GI,Mixup}     & \textbf{0.679} & \textbf{0.713} & \textbf{0.849} & \textbf{0.844} & \textbf{0.825} & \textbf{0.924} \\ \hline
Adv Tr SL~\cite{luc2016semantic}            & 0.417 & 0.507 & 0.698 & 0.731 & 0.753 & 0.891 \\ \hline
Adv Tr SSL~\cite{zhang2017deep}         & 0.409 & 0.506 & 0.692 & 0.692 & 0.719 & 0.874 \\ \hline
\end{tabular}
\\
\caption{Average Dice score (DSC) results over 15 runs of 20 test subjects for the proposed method and relevant works. $\ast,\dagger,\star$ denotes statistical significance over Aug\textsubscript{A,RD}, Aug\textsubscript{A,RI} abd Aug\textsubscript{A,Mixup} respectively. (Wilcoxon signed rank test with threshold p value of 0.05).}
% previous names of rows: No aug, With aug (Baseline), With ED, DFGAN, With CBA, AIFGAN, DFGAN and AIFGAN, MixUp~\cite{zhang2017mixup}, MixUp on DFGAN and AIFGANs, Adv Tr SL~\cite{luc2016semantic}, Adv Tr SSL~\cite{zhang2017deep}
\label{table:acdc_results}
\end{table*}

% ================================================================
% qualitative results - seg masks
% ================================================================
\setlength{\belowcaptionskip}{-15pt}

% ================================================================
% qualitative results - cganDeform
% ================================================================
\setlength{\belowcaptionskip}{-15pt}

% ================================================================
% Conclusion
% ================================================================
\vspace{-0.1cm}
\section{Conclusion}
\vspace{-0.1cm}
%Deployment of deep learning methods for medical image analyses in clinical setting is hindered, in part, due to the difficulty in assembling large-scale annotated datasets.
One of the challenging requirements for the success of deep learning methods in medical image analysis problems is that of assembling large-scale annotated datasets.
In this work, we propose a semi-supervised and task-driven data augmentation approach to tackle the problem of robust image segmentation in the setting of training datasets consisting of as few as 1 to 3 labelled 3d images.
This is achieved via two novel contributions: (i) learning conditional generative models of mappings between labelled and unlabelled images, in a formulation that also readily enables the corresponding segmentation annotations to be appropriately mapped and (ii) guiding these generative models with task-dependent losses.
In the small labelled data setting, for the task of segmenting cardiac MRIs, we show that the proposed augmentation method substantially outperforms the conventional data augmentation techniques. 
Interestingly, we observe that in order to obtain improved segmentation performance, the generated augmentation images do not necessarily need to be visually hyper-realistic.
%--line justifying the use of adversarial loss--.
%We devise conditional generative models to synthesize image-label pairs for data augmentation, that model two primary factors of variation: shape and intensity variations across a population.
%using deformation and additive intensity fields respectively such that they are helpful for the training of segmentation task.
%We also leverage unlabelled data which are readily available in the framework to model the stated variations inherently present in the dataset. 
%The nature of generated images leads to interesting questions about the best augmentation alternative to get maximal segmentation performance possible in a limited data setting. This task-driven approach is equally applicable to address the classification problems. 
%Future work : 
% \begin{itemize}
%     \item Extend to 3D
%     \item Domain adaptation
%     \item Explore other loss metrics
%     \item 1 model for multiple organs
% \end{itemize}
% ================================================================
% references

%
%\section*{Acknowledgments}
%We would like to thank SDSC and CRPP for the funding. We also thank NVIDIA for their GPU donation.

%\bibliographystyle{splncs_srt}
\bibliographystyle{splncs04}
\bibliography{main}
\end{document}